\documentclass[sigconf]{acmart}

\usepackage{booktabs} 
\usepackage{mathrsfs}

\copyrightyear{2017}
\acmYear{2017}
\setcopyright{acmcopyright}
\acmConference{MM '17}{October 23--27, 2017}{Mountain View, CA, USA}\acmPrice{15.00}\acmDOI{10.1145/3123266.3123286}
\acmISBN{978-1-4503-4906-2/17/10}
\begin{document}
\title{Fast Deep Matting for Portrait Animation on Mobile Phone}
\author{Bingke Zhu$^{1,2}$, Yingying Chen$^{1,2}$, Jinqiao Wang$^{1,2}$, Si Liu$^{2,3}$, Bo Zhang$^{4}$, Ming Tang$^{1,2}$}
\affiliation{
    \institution{$^1$National Lab of Pattern Recognition, Institute of Automation, Chinese Academy of Sciences, Beijing, China}
    \institution{$^2$University of Chinese Academy of Sciences, Beijing, China}
    \institution{$^3$Institute of Information Engineering, Chinese Academy of Sciences, Beijing, China}
    \institution{$^4$North China University of Technology, Beijing, China}
    \institution{\{bingke.zhu, yingying.chen, jqwang, tangm\}@nlpr.ia.ac.cn}
    \institution{liusi@iie.ac.cn, zhangbo@ncut.edu.cn}
}

\begin{abstract}
Image matting plays an important role in image and video editing. However, the formulation of image matting is inherently ill-posed. Traditional methods usually employ interaction to deal with the image matting problem with trimaps and strokes, and cannot run on the mobile phone in real-time. In this paper, we propose a real-time automatic deep matting approach for mobile devices. By leveraging the densely connected blocks and the dilated convolution, a light full convolutional network is designed to predict a coarse binary mask for portrait image. And a feathering block, which is edge-preserving and matting adaptive, is further developed to learn the guided filter and transform the binary mask into alpha matte. Finally, an automatic portrait animation system based on fast deep matting is built on mobile devices, which does not need any interaction and can realize real-time matting with 15 fps. The experiments show that the proposed approach achieves comparable results with the state-of-the-art matting solvers.
\end{abstract}


\maketitle

\section{Introduction}

\begin{figure}
\centering
\includegraphics[height=4.5cm,width=8.7cm]{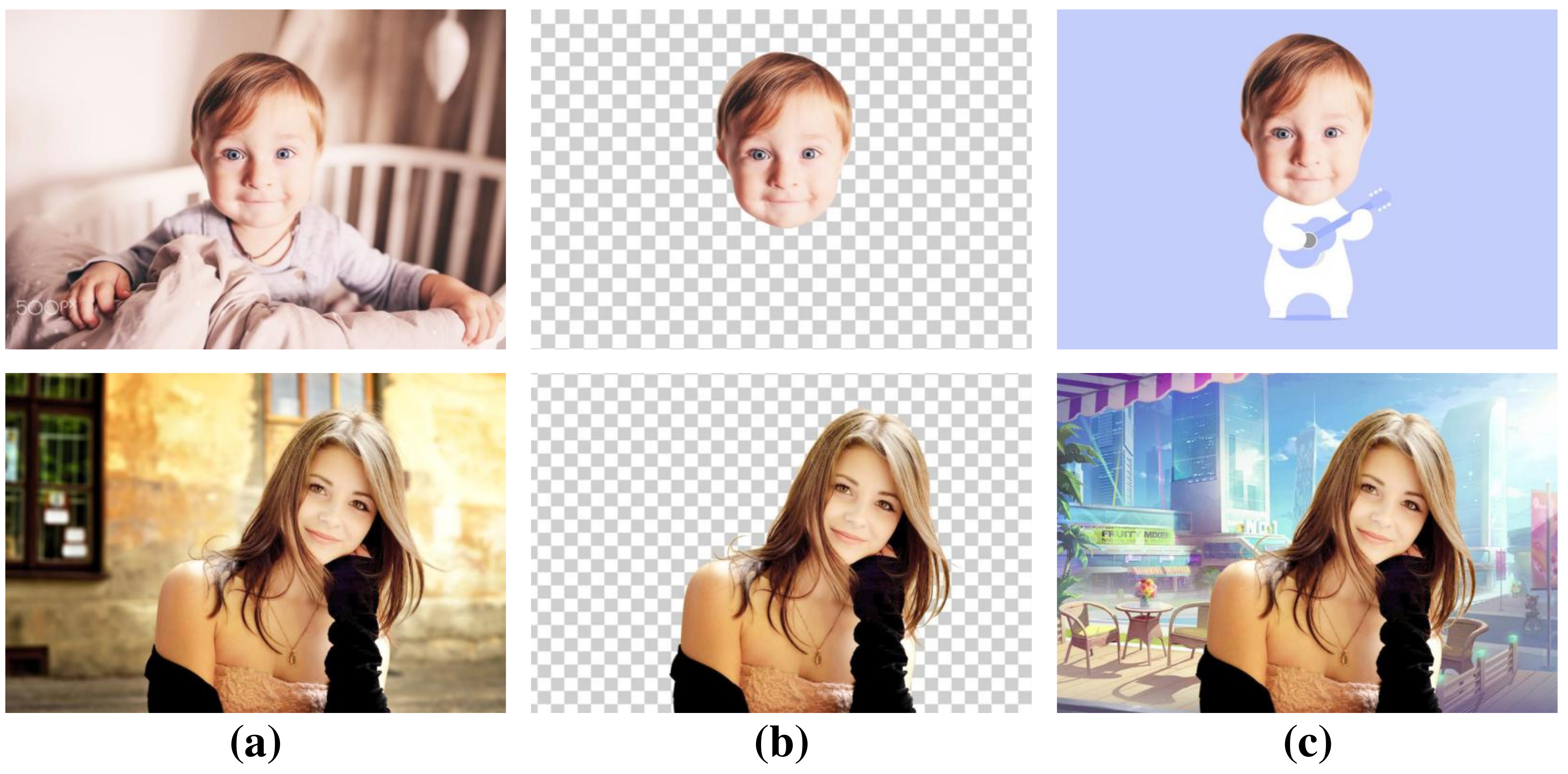}
\caption{Examples of portrait animation. (a) The original images which are the inputs to our system.
(b) The foregrounds of the original images, which are computed by the Eq. (\ref{equ:I_alpha}).
(c) The portrait animation on mobile phone.}
\label{tag:portraitanimation}
\end{figure}

Image matting plays an important role in computer vision, which has a number of applications, such as virtual reality, augmented reality, interactive image editing, and image stylization \cite {shen2016automatic}. As people are increasingly taking selfies and uploading the edited to the social network with mobile phones, real-time matting technique is in demand which can handle real world scenes. However, the formulation of image matting is still an ill-posed problem. Given an input image $I$, image matting problem is equivalent to decomposing it into foreground $F$ and background $B$ in assumption that $I$ is blended linearly by $F$ and $B$:
\begin{equation}
  I = \alpha F + (1 - \alpha )B,
  \label{equ:I_alpha}
\end{equation}
where $\alpha$ is used to evaluate the foreground opacity(alpha matte). In natural image matting, all quantities on the right-hand side of the composting Eq.~(\ref{equ:I_alpha}) are unknown, which makes the original matting problem ill-posed.

\begin{figure*}
\includegraphics[height=3.9cm]{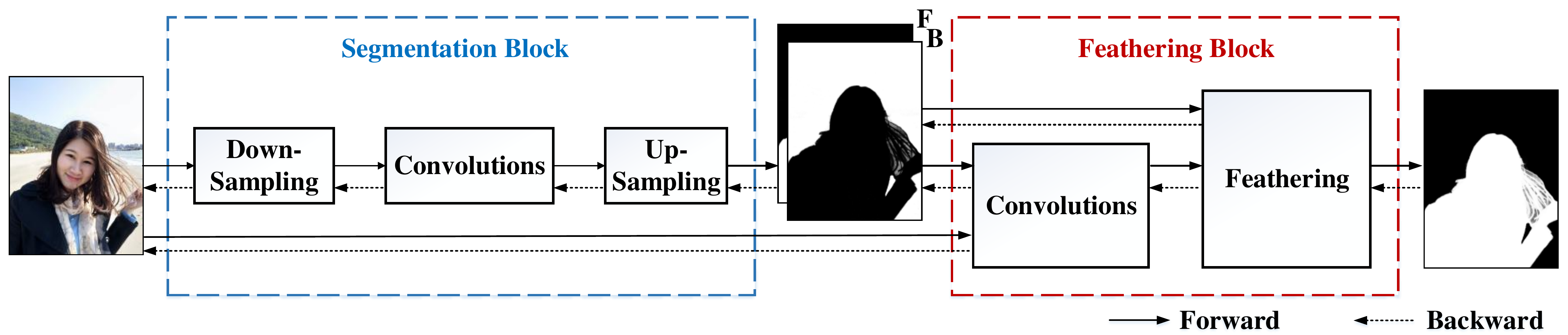}
\caption{Pipeline of our end-to-end real-time image matting framework. It includes a segmentation block and a feathering block.}
\label{tab:overallpipeline}
\end{figure*}

To alleviate the difficulty of this problem, popular image matting techniques such as \cite{levin2008closed,he2010fast,gastal2010shared,shen2016deep} require users to specify foreground and background color samples with trimaps or strokes, which makes a rough distinction among foreground, background and regions with unknown opacity. Benefit from the trimaps and strokes, the interactive mattings have high accuracy and have the capacity to segment the elaborate hair. However, despite the high accuracy of these matting techniques, these methods rely on user involvement (label the trimaps or strokes) and require extensive computation, which restricts the application scenario. In order to deal with the interactive problem, Shen et al. \cite{shen2016deep} proposed an automatic matting with the help of semantic segmentation technique. But their automatic matting have a very high  computational complexity and it  is still slow even on the GPU.

In this paper, to solve user involvement and apply matting technique on mobile phones, we propose a real-time automatic matting method on mobile phone with a segmentation block and a feathering block. Firstly, we calculate a coarse binary mask by a light full convolutional network with dense connections. And then we design a learnable guided filter to obtain the final alpha matte through a feathering block. Finally, the predicted alpha matte is calculated through a linear transformation of the coarse binary mask, whose weights are obtained from the learnt guided filter. The experiments show that our real-time automatic matting method achieves comparable results to the general matting solvers. On the other hand, we demonstrate that our system possesses adaptive matting capacity dividing the parts of head and the neck precisely, which is unprecedented in other alpha matting methods. Thus, our system is useful in many applications like real-time video matting on mobile phone. And real-time portrait animation on mobile phone can be realized using our fast deep matting technique as shown in some examples in Figure \ref{tag:portraitanimation}. Compared with existing methods, the major contributions of our work are three-fold:

1. A fast deep matting network based on segmentation block and feathering block is proposed for mobile devices. By adding the dialed convolution into the dense block structure, we propose a light dense network for portrait segmentation.

2. A feathering block is proposed to learn the guided filter and transform the binary mask into alpha matte, which is edge-preserving and possesses matting adaptive capacity.

3. The proposed approach can realize real-time matting on the mobile phones and achieve comparable results with the state-of-the-art matting solvers.

\section{Related Work}
\textbf{\emph{Image Matting:}} For interactive matting, in order to infer the alpha matte in the unknown regions, a Bayesian matting \cite{chuang2001bayesian} is proposed to model background and foreground color samples as Gaussian mixtures. Levin et al. \cite{levin2008closed} developed closed matting method to find the globally optimal alpha matte. He et al. \cite{he2010fast} computed the large-kernel Laplacian to accelerate matting Laplacian computation. Sun et al. \cite{sun2004poisson} proposed Poisson image mating to solve the homogenous Laplacian matrix. Different from the methods based on the Laplacian matrix, shared matting \cite{gastal2010shared} presented the first real-time matting technique on modern GPUs by shared sampling. Benefit from the performance of deep learning, Xu et al. \cite{xu2017deep} developed an encoder-decoder network to predict alpha matte with the help of trimaps. Recently, Aksoy et al. \cite{aksoy2017designing} designed an effective inter-pixel information flow to predict the alpha matte and reach the state of the art. Different from the interactive matting, Shen et al. \cite{shen2016deep} first proposed an end-to-end convolutional network to train the automatic matting.

With the rapid development of mobile devices, automatic matting is a very useful technique for image editing. However, despite the high efficiency of these techniques, both interactive matting and automatic matting cannot reach real-time matting on CPUs and mobile phone. The Laplacian-based methods need to compute the Laplacian matrix and its inverse, which is an $N \times N$ symmetric matrix, where $N$ is the number of unknowns. Neither non-iterative methods \cite{levin2008closed}, nor iterative methods \cite{he2010fast} can solve the sparse linear system efficiently due to the high computing cost. The real-time image matting methods like shared matting can be real-time with the help of powerful GPUs, but these methods are still slow on CPUs and mobile phone. \\
\textbf{\emph{Semantic Segmentation:}} Long et al. \cite{long2015fully} first proposed to solve the problem of semantic segmentation with fully convolutional network. ENet \cite{paszke2016enet} is a light full convolutional network for semantic segmentation, which can reach real-time on GPU. Huang et al. \cite{huang2016densely} and Bengio et al. \cite{Jgou2016TheOH} proposed a dense network with several densely connected blocks, which make the receptive filed of prediction more dense. Specially, Chen et al. \cite{chen2014semantic} developed an operation called dilated convolution, which makes the receptive filed more dense and bigger. Similarly, Yu et al. \cite{yu2015multi} proposed to use dilated convolution to aggregate the context information. On the other hand, semantic and multi-scale information is important in image understanding and semantic segmentation \cite{Wang2008AMS,Wang2014BilayerST,Zhou2016SemanticUO,Handa2015SceneNetUR}. In order to absorb more semantic and multi-scale information, Zhao et al. \cite{zhao2016pyramid} proposed a pyramid structure with dilated convolution, Chen et al. \cite{Chen2017DeepLabSI} proposed an atrous spatial pyramid pooling (ASPP), and Peng et al. \cite{Peng2017LargeKM} used the large kernel convolutions to extract features. Based on these observations, we try to combine the works of Huang \cite{huang2016densely} and Chen \cite{chen2014semantic} to create a light dense network,which can predict the segmentation densely in several layers. \\
\textbf{\emph{Guided Filter:}} Image filtering plays an important role in computer vision, especially after the development of convolutional neural network. He et al. \cite{he2013guided} proposed an edge-preserving filter - guided filter to deal with the problem of gradient drifts, which is a common problem in image filtering. Despite the powerful performance of convolutional neural network, it loses the information of edge when it comes to pixel-wise predicted assignments such as semantic segmentation and alpha matting due to its piecewise smoothing.
\begin{figure}
\includegraphics[height=12.5cm,width=9.0cm]{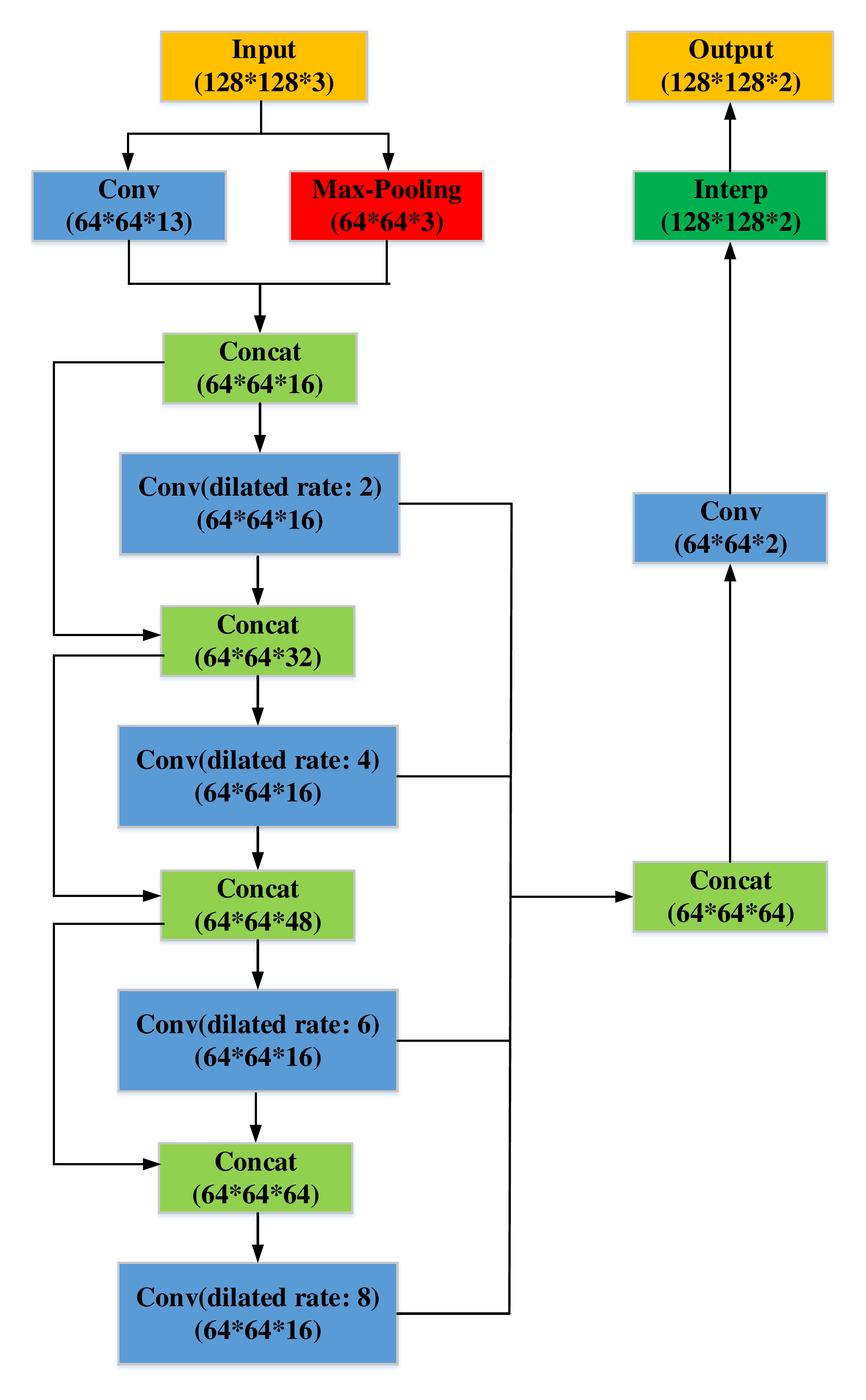}
\caption{Diagram of our light dense network for portrait segmentation.}
\label{tab:LDN}
\end{figure}
\section{Overview}
The pipeline of our system is presented in Figure \ref{tab:overallpipeline}. The input is a color image $I$ and the output is the alpha matte $\mathscr{A}$. The network consists of two stages. The first stage is a portrait segmentation network which takes an image as input and obtains a coarse binary mask. The second stage is a feathering module that refines the foreground/background mask to the final alpha matte. The first stage provides a coarse binary mask in a fast speed with a light full convolutional network, while the second stage refines the coarse binary mask with a single filter, which reduces the error greatly. We will describe our algorithm with more details in the following sections.
\subsection{Segmentation Block}
In order to segment the foreground with a fast speed, we propose a light dense network in the segmentation block. The architecture of the light dense network is presented in Figure \ref{tab:LDN}. Output sizes are reported for an example input image resolution of $128 \times 128$. The initial block is a concatenation of a $3 \times 3$ convolution and a max-pooling, which is used to down sample the input image. The dilated dense block contains four convolutional layers in different dilated rates and four densely connections. The concatenation of four convolutional layers are sent to the final convolution to obtain a binary feature maps. Finally, we interpolate the feature maps to get the score maps which have the same size as the original image. Specifically, our light dense network has 6 convolutional layers and 1 max-pooling layer.

The motivation of this architecture is inspired by several popular full convolutional networks for semantic segmentation. Inspired by the real-time segmentation architecture of ENet \cite{paszke2016enet} and Inception network \cite{szegedy2016inception}, we use the initial block in ENet to down sample the input, which can maintain more information than that of max-pooling and bring less computation cost than that of convolution. Then the down-sampling maps are sent to the dilated dense block, which is inspired by the densely connected convolution network \cite{huang2016densely} and the dilated convolution network \cite{chen2014semantic}. Each layer of the dilated dense block obtains different field of view, which is important in the multi-scale semantic segmentation. The dilated dense block can seize the foreground in variable sizes, thus obtain a better segmentation mask by the final classifier.
\subsection{Feathering Block}
The segmentation block can produce a coarse binary mask to the alpha matte. However, the binary mask cannot represent the alpha matte due to the coarse edge. Despite fully convolution networks have been proved to be effective in the semantic segmentation \cite{long2015fully}, the methods like \cite{xu2017deep} using fully convolutional networks may suffer from the gradient drifts because of the pixel-wise smoothing in the convolutional operations. In this section, we will discuss a feathering block to refine the binary mask and solve the problem of gradient drifts.

\subsubsection{Architecture}
The convolutional operations play important roles in the classification networks. However, alpha matting is a pixel-wise predicted task, and it is unsuitable to predict the edge of the objects with the convolutional operations due to its piecewise smoothing. The state of the art method \cite{chen2014semantic} relieves this problem using the conditional random field (CRF) model to refine the edge of the objects, but it costs too much computing memory and it cannot solve the problem in the convolutional neural networks intrinsically. Motivated by the gradient drifts of the convolution operations, we developed a sub-network to learn the filters of the coarse binary mask, which does not suffer from the gradient drifts.

\begin{figure}
\includegraphics[height=9.8cm]{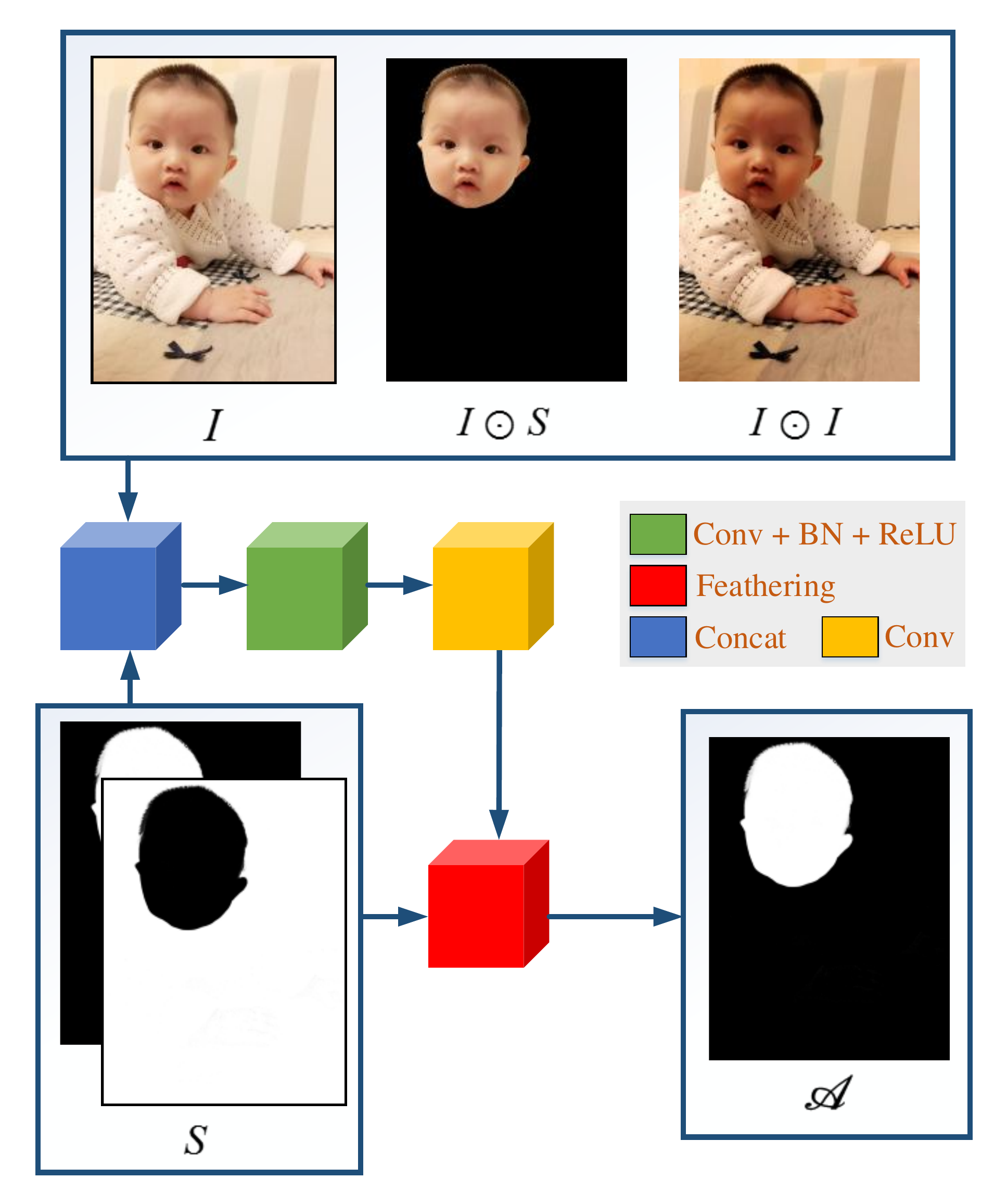}
\caption{Architecture of the feathering block, which is used to learn the filter refining the coarse binary mask. $I$ is the input image, $S$ is the score maps output from the segmentation network, $\mathscr{A}$ is the alpha matte which is the final output of the whole network, and $\odot $ is the hadamard product.}
\label{tab:FB}
\end{figure}

The architecture of the feathering block is presented in Figure \ref{tab:FB}. The inputs to the feathering block are an image, the corresponding coarse binary mask, the square of the image as well as the product of the image and its binary mask. The design of these inputs is inspired by the guided filter \cite{he2013guided}, whose weights are designed as a function of these inputs. We concatenate the inputs and send the concatenation into the convolutional network which contains two $3 \times 3$ convolutional layers. Then we can obtain three maps corresponding to the weights and bias of the binary mask. The feathering layer can be represented as a linear transform of the coarse binary mask in sliding windows centered at each pixel:
\begin{equation}
 {\alpha _i} = {a_k}{S_{Fi}} + {b_k}{S_{Bi}} + {c_k},\forall i \in {\omega _k},
 \label{equ:alphai}
\end{equation}
where $\alpha$ is the output of feathering layer represented as alpha matte, ${S_F}$ is the foreground score from the coarse binary mask, ${S_B}$ is the background score, $i$ is the location of the pixel, and $\left( {{a_k},{b_k},{c_k}} \right)$ are the linear coefficients assumed to be constant in the $k$-th sliding window $\omega _k$. Thus, we have:
\begin{equation}
 {q_i} = {a_k}{F_i} + {b_k}{B_i} + {c_k}{I_i},\forall i \in {\omega _k},
 \label{equ:qi}
\end{equation}
where ${q_i} = {\alpha _i}{I_i}$, ${F_i} = {I_i}{S_{Fi}}$, ${B_i} = {I_i}{S_{Bi}}$, and $I$ is the input image.
\begin{figure*}
\includegraphics[height=6.5cm, width=15.7cm]{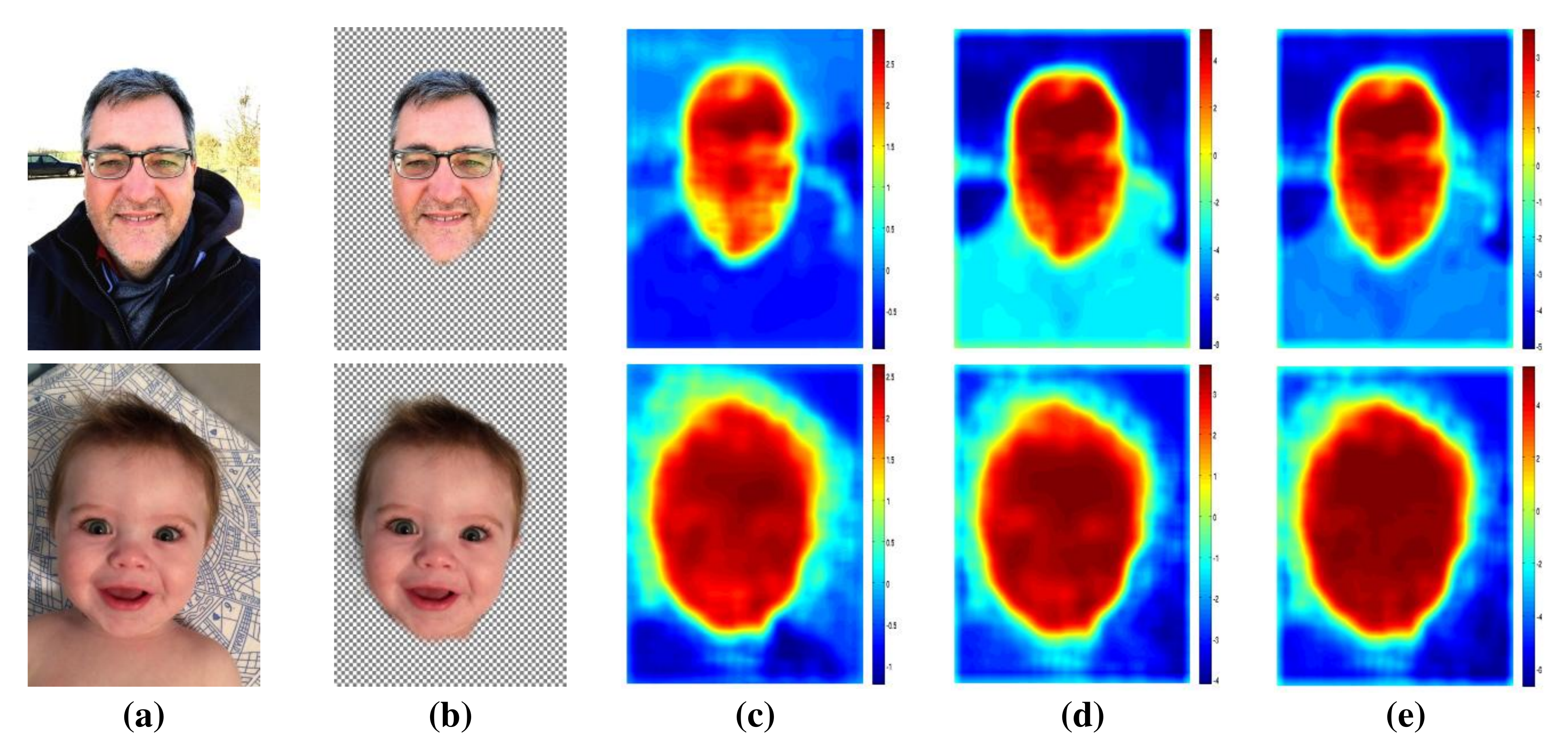}
\caption{Examples of the filters learnt from the feathering block. (a) The input images. (b) The foregrounds of the original images, which are calculated by the Eq.~(\ref{equ:I_alpha}). (c) The weights ${a_k}$ of feathering block in Eq.~(\ref{equ:alphai}) and Eq.~(\ref{equ:qi}). (d) The weights ${b_k}$ of feathering block in Eq.~(\ref{equ:alphai}) and Eq.~(\ref{equ:qi}). (e) The weights ${c_k}$ of feathering block in Eq.~(\ref{equ:alphai}) and Eq.~(\ref{equ:qi}).}
\label{tab:FBweights}
\end{figure*}

From (3) we can get the derivative
\begin{equation}
 \nabla q = a\nabla F + b\nabla B + c\nabla I,
 \label{equ:proofqFBI}
\end{equation}
where $q = \alpha I$, $F = I{S_F}$, $B = I{S_B}$. It ensures that the feathering block possesses the property of edge-preserving and matting adaptive. As discussed above, both two score maps ${S_F}$ and ${S_B}$ would have strong responses in the area of edge because of the uncertainty in these areas. However, it is worth to note that the score maps of foreground and background are allowed to have inaccurate responses when the parameters $a$, $b$, $c$ are trained well. In this case, we hope that the parameters $a$, $b$ become as small as possible, which means that the inaccurate responses are suppressed. In other word, the feathering block can preserve the edge as long as the absolute values of $a$, $b$ are set to small in the area of edge while $c$ is predominant on it. Similarly, if we want to segment the neck apart from the head, the parameters $a$, $b$, $c$ should be set to small in the area of neck. Moreover, it is interesting that the feathering block performs like ensemble learning because we can treat $F$, $B$, $I$ as the classifiers and the parameters $a$, $b$, $c$ as the weights for classification.

When we apply the linear model to all sliding windows in the entire image, the value ${\alpha _i}$ is not the same in different windows. We leverage the same strategy as He et al. \cite{he2013guided}. After computing $\left( {{a_k},{b_k},{c_k}} \right)$ for all sliding windows in the image, we average all the possible values of ${\alpha _i}$:
\begin{equation}
\begin{aligned}
\alpha _i&=\frac{1}{\left| w \right|}\sum_{k:i\in w_k}{a_kS_{Fi}+b_kS_{Bi}+c_k}\\
&=\overline{a}_iS_{Fi}+\overline{b}_iS_{Bi}+\overline{c}_i,
\end{aligned}
\end{equation}
where
\begin{displaymath}
\overline{a}_i=\frac{1}{\left| \omega \right|}\sum_{k\in {\omega_i}}{a_k}, \overline{b}_i=\frac{1}{\left| \omega \right|}\sum_{k\in {\omega_i}}{b_k}, \overline{c}_i=\frac{1}{\left| \omega \right|}\sum_{k\in {\omega_i}}{c_k}.
\end{displaymath}
With this modification, we can still have $\nabla q\approx \overline{a}\nabla F+\overline{b}\nabla B+\overline{c}\nabla I$ because $\left( {{{\overline a }_k},{{\overline b }_k},{{\overline c }_k}} \right)$ are the average of the filters, and their gradients should be much smaller than that of $I$ near strong edges.

To determine the linear coefficients, we design a sub-network to seek the solution. Specially, our network leverages a loss function including two parts, which makes the alpha predictions more accurate. The first loss ${L_\alpha }$ measures the alpha matte, which is the absolute difference between the ground truth alpha values and the predicted alpha values at each pixel. And the second loss $L_{color}$ is a compositional loss, which is the L2-norm loss function for the predicted RGB foreground. Thus, we minimize the following cost function:
\begin{equation}
L = {L_\alpha } + {L_{color}},
\end{equation}
where
\begin{displaymath}
\begin{aligned}
&L_\alpha ^i = \sqrt {{{\left( {\alpha _{gt}^i - \alpha _p^i} \right)}^2} + {\varepsilon ^2}}, \\
&L_{color}^i = \sum\limits_{j \in \left\{ {R,G,B} \right\}} {\sqrt {{{\left( {\alpha _{gt}^iI_j^i - q_j^i} \right)}^2} + {\varepsilon ^2}} } , \\
&\alpha _p^i = {a_k}{S_{Fi}} + {b_k}{S_{Bi}} + {c_k}.
 \end{aligned}
\end{displaymath}

This loss function is chosen for two reasons. Intuitively, we hope to obtain an accurate alpha matte through the feathering block in the end, thus we use the alpha matte loss ${L_\alpha }$ to learn the parameters. On the other hand, the second loss ${L_{color}}$ is used to maintain the information of the input image as much as possible. Since the score maps ${S_F}$ and ${S_B}$ would lose the information of edges due to the pixel-wise smoothing, we need to use the input image to recover the lost detail of the edge. Therefore, we leverage the second loss ${L_{color}}$ to guide the learning process. Despite existing deep learning methods like \cite{Cho2016Natural,shen2016deep,xu2017deep} used the similar loss function to solve the alpha matting problem, we have a totally different motivation and a different solution. Their loss functions are used to estimate the expression of alpha matting without considering the gradient drifts for the convolution operation, which causes the pixel-wise smooth alpha matte and loses the gradients of input image, while our loss function is edge-preserving. Through the gradient propagation, the parameters of feathering layer can be updated pixel-wise so as to correct the edges. It is because that the guided filter is an edge-preserving filter. However, other deep learning methods can only update the parameters in the fixed windows. As a result, their approaches would make the results pixel-wise smooth, while our results are edge-preserving.

\subsubsection{Guided Image Filter}
The guided filter \cite{he2013guided} proposed a local linear model:
\begin{equation}
{q_i} = {a_k}{I_i} + {b_k},\forall i \in {\omega _k},
\end{equation}
where $q$ is a linear transform of $I$ in a window ${\omega _k}$ centered at the pixel $k$, and $\left( {{a_k},{b_k}} \right)$ are some linear coefficients assumed to be constant in ${\omega _k}$. In order to determine the linear coefficients, they minimize the following cost function in the window:
\begin{equation}
E\left( {{a_k},{b_k}} \right) = \sum\limits_{i \in {\omega _k}} {\left( {{{\left( {{a_k}{I_i} + {b_k} - {p_i}} \right)}^2} + \varepsilon a_k^2} \right)}.
\end{equation}

Our feathering block does not use the same form like [10] because we hope to obtain not only an edge-preserving filter, but also a filter with the capacity of matting adaptive. It means the filters can suppress the inaccurate response and obtain a finer matting result. As a result, the guided filter needs to rely on an accurate binary mask while our filter does not. Therefore, the feathering block can be viewed as an extension of the guided filter. We extend the linear transform of the image to the linear transform of the foreground image, the background image as well as the whole image. Moreover, instead of solving the linear regression to obtain the linear coefficients, we leverage a sub-network to optimize the cost function, which should be much faster and reduce the computational complexity.

Figure \ref{tab:FBweights} shows that our feathering block is closely to the guided filter in \cite{he2013guided}. Both the filter of ours and guided filter have high response on the high variance regions. However, it is obvious that there are great distinctions between our filter and guided filter. The differences between our filter and guided filter can be summarized into two aspects briefly:

1. \textit{Different inputs.} The inputs of guided filter are the image as well as the corresponding binary mask while our inputs are the image and the score maps of the foreground and the background.

2. \textit{Different outputs.} The output of guided filter is a feathering results relied on an accurate binary mask, while our learnt guided filter outputs an alpha matte or a foreground image, which possesses the fault tolerance to the binary mask.

In fact, our feathering block is matting oriented, therefore, the linear transform model of learnt guided filter in Eq. (\ref{equ:qi}) is inferred from the linear transform model of matting filter in Eq. (\ref{equ:alphai}). Thus, our learnt guided filter has multiple product terms but no constant term. Though two filters are derived in different process, they have the same properties such as edge-preserving, which has been proved in Eq. (\ref{equ:proofqFBI}).
Moreover, Figure \ref{tab:FBweights} shows that our learnable guided filter possesses adaptive matting capacity. The general matting methods like \cite{levin2008closed, gastal2010shared} would take the neck as the foreground if we define the neck in the image as unknown region, because the gradient between the face and neck is small. However, our method can divide the face and neck into the foreground and background, respectively.
\subsubsection{Attention Mechanism}
Our proposed feathering block is not only an extension of the guided filter, but also can be interpreted as an attention mechanism. Figure \ref{tab:FBweights} shows several examples of guided filters learnt from the feathering block. Intuitively, it can be interpreted as an attention mechanism, which pays different attention to various parts according to factors. Specially, from the examples in Figure \ref{tab:FBweights}, we can infer that the factor $\overline a $ pays more attention to the part of object's body, the factor $\overline b $ pays more attention to the part of background, and the factor $\overline c $ pays more attention to the part of object's head. Consequently, it can be inferred that the factors $\overline a $ and $\overline b $ emphasize the matting problem locally while the factor $\overline c $ considers the matting problem globally.

\section{Experiments}
\subsection{Dataset and Implementation}
\textbf{\emph{Dataset:}} We collect the primary dataset from \cite{shen2016deep}, which is collected from Flickr. After training on this dataset, we release our app to obtain more data for training. Furthermore, we hired tens of well-trained students to accomplish the annotation work. All the ground truth alpha mattes are labelled with KNN matting \cite{chen2013knn} firstly, and refined carefully with Photoshop quick selection.

Specially, we labelled the alpha matte in two ways. The first way of labelling is same as \cite{shen2016deep, xu2017deep}, which labels the whole human as the foreground. The second way of labelling only labels the heads of human as the foreground. The head labels help us demonstrate that our proposed method possesses adaptive matting capacity dividing the parts of head and neck precisely. After labelling process, we collect 2,000 images with high-quality mattes, and split them randomly into training and testing sets with 1,800 and 200 images respectively.

For data augmentation, we adopt random mirror and random resize between 0.75 and 1.5 for all images, and additionally add random rotation between -30 and 30 degrees, and random Gaussian blur. This comprehensive data augmentation scheme prevents the network overfitting and greatly improves the performance of our system to handle new images with possibly different scale, rotation, noise and intensity. \\
\textbf{\emph{Implementation details:}} We setup our model training and testing experiments on Caffe platform \cite{jia2014caffe}. With the model illustrated in Figure \ref{tab:overallpipeline}, we use a stochastic gradient descent (SGD) solver with cross-entropy loss function, batch size 256, momentum 0.99 and weight decay 0.0005. We train our model without any further post-processing module nor pre-training. The unknown weights are initialized with random values. Specially, we train our network with a three-stage strategy. Firstly, we train the light dense network with a learning rate of 1e-3 and 1e-4 in the first 10k and the last 10k iterations, respectively. Secondly, the weights in light dense network is fixed and we only train the feathering block with a learning rate set to 1e-6 which will be divided by 10 after 10k iterations. Finally, we train the whole network with a fixed learning rate of 1e-7 for 20k iterations. We found that the three-stage training strategy makes the training process more stable. All experiments on computer are performed on a system of Core E5-2660 @2.60GHz CPU and a single NVIDIA GeForce GTX TITAN X GPU with 12GB memory. We also test our approach on the mobile phone with a Qualcomm Snapdragon 820 MSM8996 CPU and Adreno 530 GPU.

\subsection{Head Matting Dataset}

\begin{figure*}
\includegraphics[height=8.5cm, width=15.3cm]{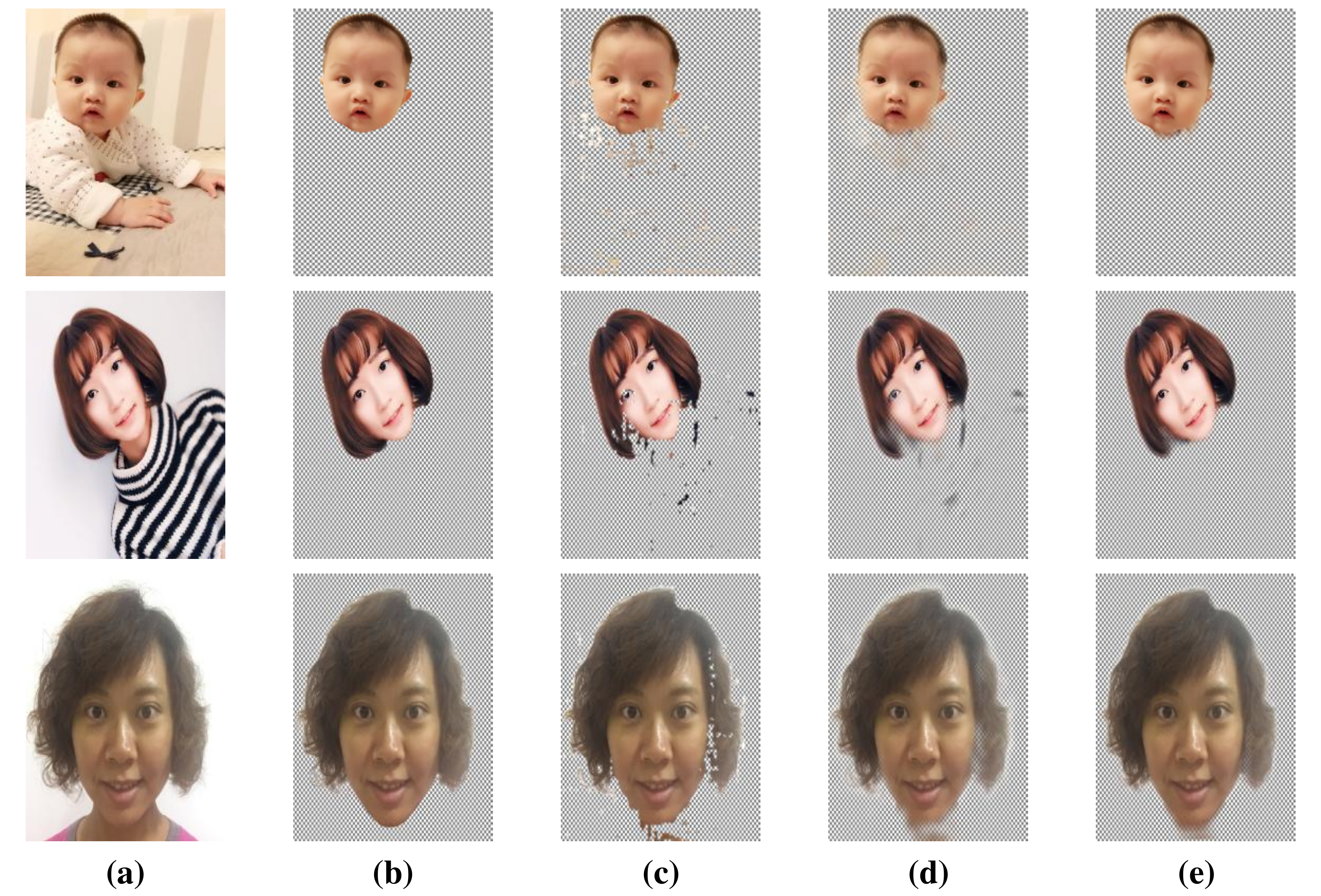}
\caption{Several results from binary mask, binary mask with guided filter, and alpha matte from our feathering block. (a) The original images. (b) The ground truth foreground. (c) The foreground calculated by the binary mask. (d) The foreground calculated by the binary mask with guided filter. (e) The foreground calculated by the binary mask with our feathering block.}
\label{tag:comparemethods}
\end{figure*}
\textbf{Accuracy Measure.} We select the gradient error and mean squared error to measure matting quality, which can be expressed as:
\begin{equation}
 G\left( {\mathscr{A},{\mathscr{A}^{gt}}} \right) = \frac{1}{K}\sum\limits_i {\left\| {\nabla {\mathscr{A}_i} - \nabla \mathscr{A}_i^{gt}} \right\|},
\end{equation}
\begin{equation}
 MSE\left( \mathscr{A},\mathscr{A}^{gt} \right) =\frac{1}{K}\sum_i{\left( \mathscr{A}-\mathscr{A}^{gt} \right) ^2},
\end{equation}
where $\mathscr{A}$ is the predicted matte and $\mathscr{A}^{gt}$ is the corresponding ground truth, $K$ is the number of pixels in $\mathscr{A}$, $\nabla $ is the operator to compute gradients.

\begin{table}[ht]
\caption{Results on Head Matting Dataset. We compare the components of our proposed system with state-of-the-art semantic segmentation networks Deeplab and PSPN \cite{chen2014semantic, zhao2016pyramid}. Light Dense Network (LDN) greatly increases the speed and Feathering Block (FB) deceases the gradient error(Grad. Error) and mean squared error(MSE) greatly in our system. Additionally, our feathering block has a better performance than guided filter (GF).}
\begin{tabular}{p{2.3cm}p{0.7cm}p{0.7cm}p{1.8cm}p{1.2cm}}
\toprule
Approaches &CPU (ms) &GPU (ms) &Grad. Error($\times 10^{-3}$) &MSE ($\times 10^{-3}$) \\ \midrule
Deeplab101 \cite{chen2014semantic} &   1243 &   73 &   3.61 & 5.84\\
PSPN101 \cite{zhao2016pyramid} &   1289 &   76 &   3.88 & 7.07 \\
PSPN50 \cite{zhao2016pyramid} &   511 &   44 &   3.92 & 8.04 \\
LDN &  \textbf{27} &  \textbf{11} &  9.14 & 27.49 \\
Deeplab101+FB & 1424 & 78 & 3.46 & 4.23 \\
PSPN101+FB & 1343 & 83 & \textbf{3.37} & \textbf{3.82} \\
PSPN50+FB & 548 & 52 & 3.64 & 4.67 \\
LDN+GF & 236 & 220 & 4.66 & 15.34 \\
\hline
\textbf{LDN+FB} & 38 & 13 & 3.85 & 7.98  \\
\bottomrule
\end{tabular}\label{tab:headdata}
\end{table}

\textbf{Method Comparison.} We compare several automatic schemes for matting and report the performance of these methods in Table~\ref{tab:headdata}. From the results, we can draw the conclusion that our proposed method \emph{LDN+FB} is efficient and effective. Firstly, comparing the portrait segmentation networks, \emph{LDN} produces a coarse binary mask, which increases 3 times gradient error and 3 to 5 times mean squared error approximately, compared to the semantic segmentation networks. However, \emph{LDN} is 20 to 60 times faster than the semantic segmentation networks on CPU. Secondly, taking our feathering block into consideration, we find that \emph{FB} can refine the portrait segmentation networks with little effort on CPU or GPU. Moreover, the feathering block decreases the gradient error and mean squared error of our proposed \emph{LDN} greatly, which decreases $30\%$ of the errors approximately. Furthermore, when compared to the guided filter, it deceases 2 times of mean squared  error. Thirdly, our proposed \emph{LDN+FB} structure have comparable accuracies with the combinations of segmentation networks and feathering blocks, while it is 15 to 40 times faster.
\begin{table}
\caption{Results on Human Matting Dataset of \cite{shen2016deep}. DAPM means the approach of Deep Automatic Portrait Matting in \cite{shen2016deep}. LDN means the light dense network and FB means the feathering block. We report our speed on computer(comp.) and mobile phone(phone.) respectively.}
\begin{tabular}{p{1.4cm}p{1cm}p{1cm}p{1cm}p{1cm}p{1.2cm}}
\toprule
Approaches &CPU /comp. (ms) &GPU /comp. (ms) &CPU /phone. (ms) &GPU /phone. (ms) &Grad. Error ($\times 10^{-3}$) \\ \midrule
DAPM & 6000 & 600 &- &- & 3.03 \\
LDN+FB & 38 & 13 &140 &62 & 7.40 \\
\bottomrule
\end{tabular}\label{tab:humandata}
\end{table}
Visually, we present the results of binary mask, binary mask with guided filter, and alpha matte from our feathering block in Figure \ref{tag:comparemethods}. It is obvious that the binary masks are very coarse which lose the edge shape and contain lots of isolated holes. The binary mask with guided filter can fill the holes thereby enhancing the performance of the mask, however, the alpha mattes are over transparent due to the coarse binary mask. In contrast, our proposed system achieves better performance by making the foreground clearer and remaining the edge shape.

In order to illustrate the efficiency and effectiveness of the light dense block and feathering block, we discuss them separably. Firstly, comparing the four portrait segmentation networks, though our light dense network has a low performance, it decreases the computing time greatly, and it can produce a coarse segmentation with 8 convolutional layers. Secondly, the feathering block increases the performances of four portrait segmentation networks. It decreases the error of our proposed light dense network greatly, which holds a comparable results with the best results and keeps a high speed. Additionally, comparing the performances between guided filter and feathering block, we can infer that feathering block outperforms the guided filter in accuracy while keeping a high speed.

\subsection{Human Matting Dataset}
Furthermore, in order to demonstrate the efficiency and effectiveness of our proposed system. We train our system on the dataset of \cite{shen2016deep}, which contains a training set and a testing set with 1,700 and 300 images respectively, and compare our result with that of their proposed methods.

We report the performance of these automatic matting methods in Table~\ref{tab:humandata}. From the reported results, we find that our proposed method increases $4.37\times 10^{-3}$ in gradient error, compared to the error reported in \cite{shen2016deep}. However, our method decreases the time cost of \emph{5962ms} on CPU and \emph{587ms} on GPU. Consequently, it can be illustrated that our method increases the matting speed greatly while keeping a comparable result with the general alpha matting methods. Moreover, the automatic matting approach proposed in \cite{shen2016deep} can not run on the CPU of mobile phone because it needs a lot of time to infer the alpha matte. In contrast, our method can not only run on the GPU of mobile phone in real-time, but also has the capacity to run on the CPU of mobile phone.
\subsection{Comparison with Fabby}
\begin{figure*}
\includegraphics[height=7.6cm, width=16.2cm]{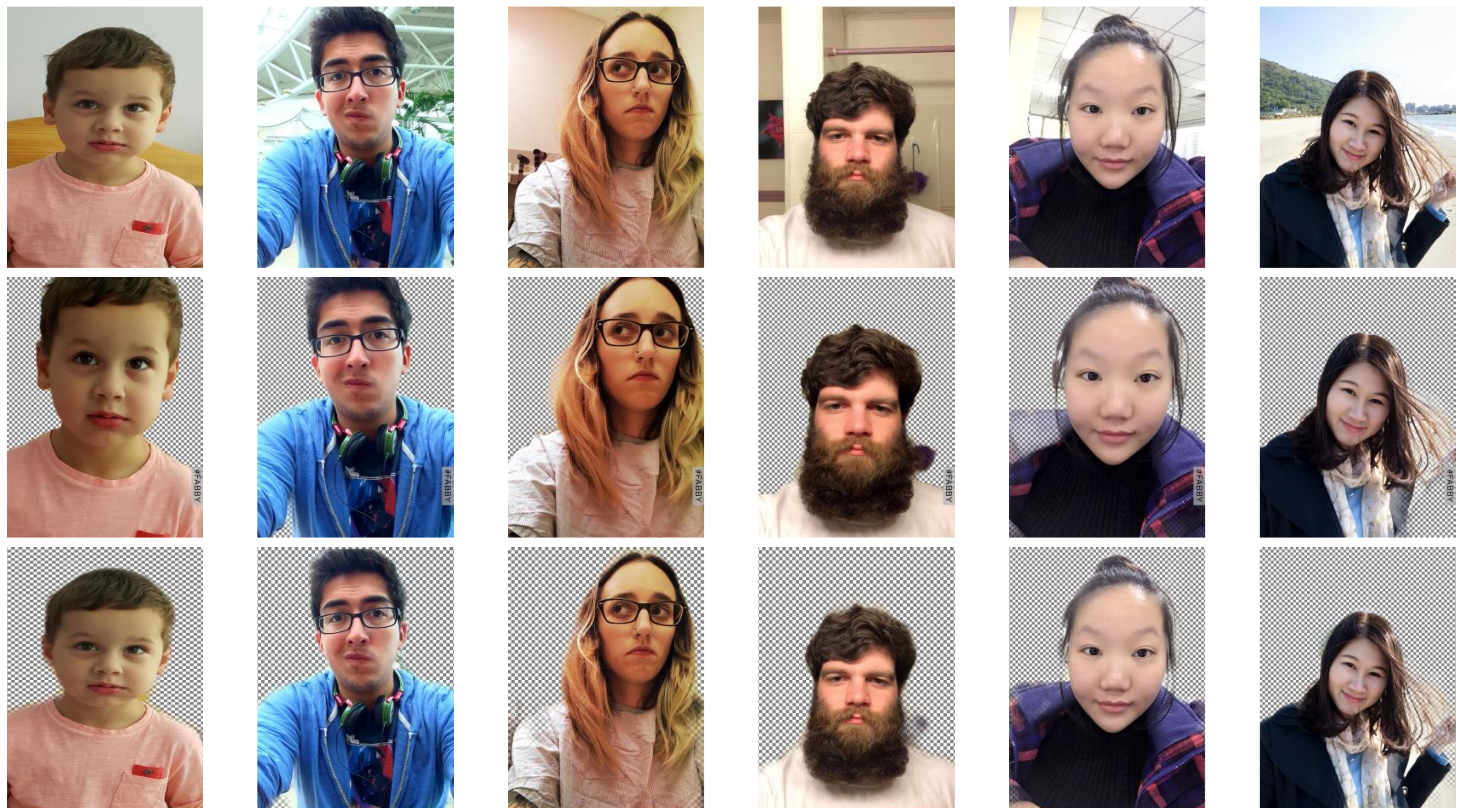}
\caption{The results between our system and real-time editing app Fabby. \textbf{First row:} The input images. \textbf{Second row:} The results from the app Fabby. \textbf{Third row:} The results from our system.}
\label{tag:compareFabby}
\end{figure*}

Additionally, we compare our system with Fabby\footnote{The app of Fabby can be downloaded on the site www.fab.by}, which is a real-time face editing app. The visual results are presented in Figure \ref{tag:compareFabby}. It is shown that we have better results than that of Fabby. Moreover, we found that Fabby cannot locate the objects such as the cases in former three columns and it fails to process some special cases like the faults in the last three columns. In contrast, our method can hold these cases well.

From the former three columns in Figure~\ref{tag:compareFabby}, we can infer that Fabby needs to detect the objects first, and then computes the alpha matte. As a result, sometimes it is unable to calculate the alpha matte of the whole object. Conversely, our method can calculate the alpha matte of all pixels in an image and thus obtain the opacity of the whole object. Furthermore, Fabby fails to divide the similar background from foreground for the case of the fourth column, and fails to segment the foreground such as collar and shoulder in last two columns, while our method outperforms Fabby in these cases. Consequently, it can be inferred that our light matting network has a more reliable alpha matte than that of Fabby.
\begin{figure}
\centering
\includegraphics[height=5.5cm,width=8.4cm]{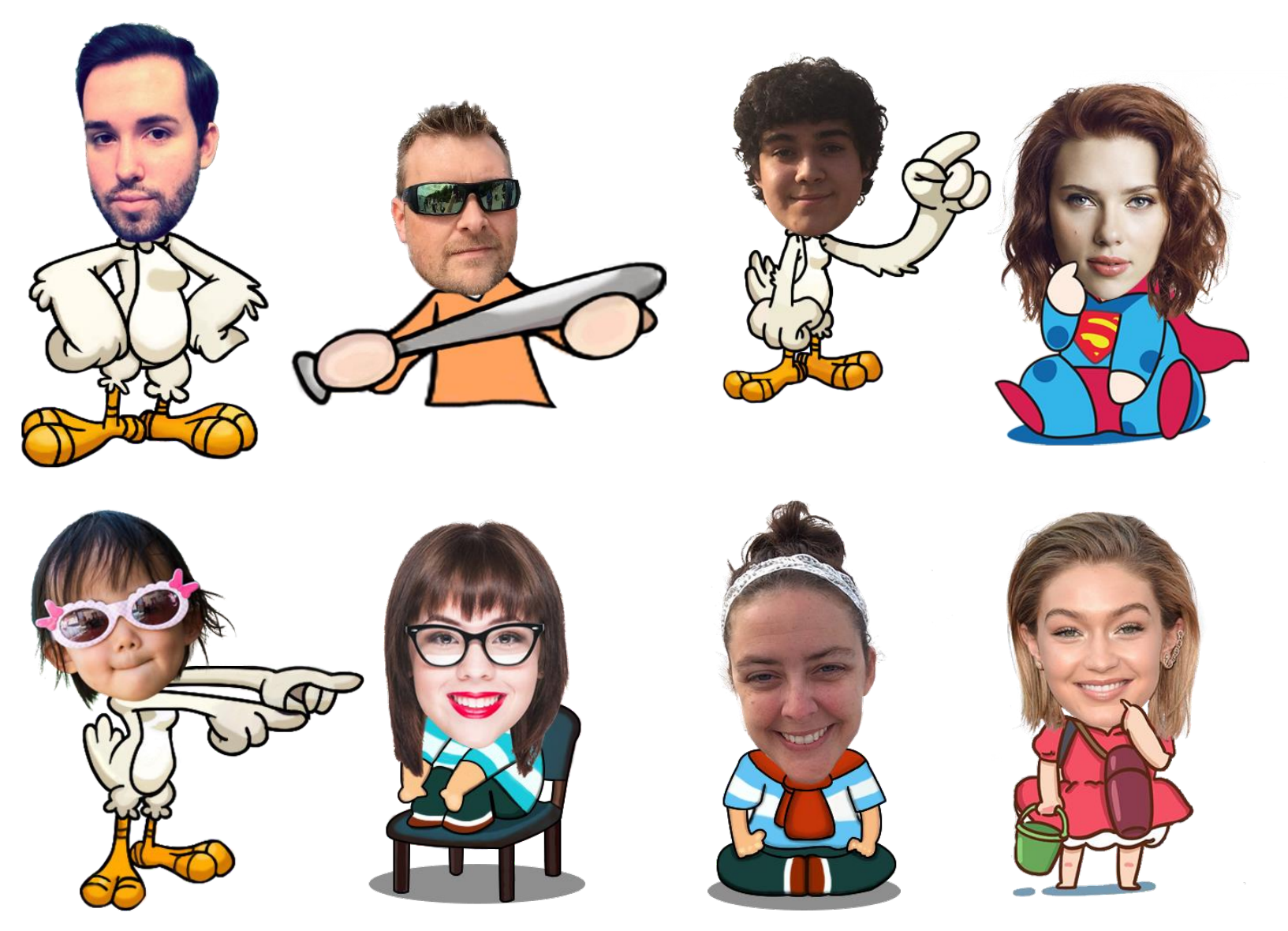}
\caption{Examples of our real-time portrait animation app.}
\label{tag:ourapp}
\end{figure}
\section{Applications}
Since our proposed system is fully automatic and can run fully real-time on mobile phone, a number of applications are enabled, such as fast stylization, color transform, background editing and depth-of-field. Moreover, it can not only process the image matting, but also real-time video matting due to its fast speed. Thus, we establish an app for real-time portrait animation on mobile phone, which is illustrated in Figure \ref{tag:ourapp}. Further, we test the speed of our method on the mobile phone with a Qualcomm Snapdragon 820 MSM8996 CPU and an Adreno 530 GPU. When tests on mobile phone, it runs 140ms on the CPU. Furthermore, after applying render script \cite{cnndroid2016} to optimize the speed with GPU in Android, our approach takes 62ms on the GPU. Consequently, we can conclude that our proposed method is efficient and effective on mobile devices.
\section{Conclusions and Future Work}
In this paper, we propose a fast and effective method for image and video matting. Our novel insight is that refining the coarse segmentation mask can be fast and effective than the general image matting methods. We developed a real-time automatic matting method with a light dense network and a feathering block. Besides, the filter learnt by the feathering block possesses edge-preserving and matting adaptive capacity. Finally, a portrait animation system based on fast deep matting is built on mobile devices. Our method does not need any interaction and can realize real-time matting on the mobile phone. The experiments show that our real-time automatic matting method achieves comparable results with the state-of-the-art matting solvers. However, our system fails to distinguish tiny details in the hair areas because we have downsampled the input image. The downsampling operation would provide a more complete result but lose the detailed information. We treat this as the limitation of our method. In future, we will try to improve our method for higher accuracy by combining multiple technologies like object detection \cite{Ren2015FasterRT,Chen2016AUM,Liu2016SSDSS,Redmon2016YOLO9000BF}, image retargeting \cite{Wang2016AdaptiveCC,Qi2016CASAIRCA} and face detection \cite{Schroff2015FaceNetAU,Qin2016JointTO,Hu2016FindingTF}.

\bibliographystyle{ACM-Reference-Format}
\bibliography{fast_deep_matting_bibliography}


\begin{thebibliography}{00}


\ifx \showCODEN    \undefined \def \showCODEN     #1{\unskip}     \fi
\ifx \showDOI      \undefined \def \showDOI       #1{#1}\fi
\ifx \showISBNx    \undefined \def \showISBNx     #1{\unskip}     \fi
\ifx \showISBNxiii \undefined \def \showISBNxiii  #1{\unskip}     \fi
\ifx \showISSN     \undefined \def \showISSN      #1{\unskip}     \fi
\ifx \showLCCN     \undefined \def \showLCCN      #1{\unskip}     \fi
\ifx \shownote     \undefined \def \shownote      #1{#1}          \fi
\ifx \showarticletitle \undefined \def \showarticletitle #1{#1}   \fi
\ifx \showURL      \undefined \def \showURL       {\relax}        \fi
\providecommand\bibfield[2]{#2}
\providecommand\bibinfo[2]{#2}
\providecommand\natexlab[1]{#1}
\providecommand\showeprint[2][]{arXiv:#2}

\bibitem[\protect\citeauthoryear{Aksoy, Ayd{\i}n, and Pollefeys}{Aksoy
  et~al\mbox{.}}{2017}]%
        {aksoy2017designing}
\bibfield{author}{\bibinfo{person}{Yag{\i}z Aksoy},
  \bibinfo{person}{Tun{\c{c}}~Ozan Ayd{\i}n}, {and} \bibinfo{person}{Marc
  Pollefeys}.} \bibinfo{year}{2017}\natexlab{}.
\newblock \showarticletitle{Designing Effective Inter-Pixel Information Flow
  for Natural Image Matting}. In \bibinfo{booktitle}{{\em Computer Vision and
  Pattern Recognition (CVPR), 2017}}.
\newblock


\bibitem[\protect\citeauthoryear{Chen, Papandreou, Kokkinos, Murphy, and
  Yuille}{Chen et~al\mbox{.}}{2015}]%
        {chen2014semantic}
\bibfield{author}{\bibinfo{person}{Liang-Chieh Chen}, \bibinfo{person}{George
  Papandreou}, \bibinfo{person}{Iasonas Kokkinos}, \bibinfo{person}{Kevin
  Murphy}, {and} \bibinfo{person}{Alan~L Yuille}.}
  \bibinfo{year}{2015}\natexlab{}.
\newblock \showarticletitle{Semantic image segmentation with deep convolutional
  nets and fully connected crfs}. In \bibinfo{booktitle}{{\em International
  Conference on Learning Representations (ICLR), 2015}}.
\newblock


\bibitem[\protect\citeauthoryear{Chen, Papandreou, Kokkinos, Murphy, and
  Yuille}{Chen et~al\mbox{.}}{2017}]%
        {Chen2017DeepLabSI}
\bibfield{author}{\bibinfo{person}{Liang-Chieh Chen}, \bibinfo{person}{George
  Papandreou}, \bibinfo{person}{Iasonas Kokkinos}, \bibinfo{person}{Kevin
  Murphy}, {and} \bibinfo{person}{Alan~L. Yuille}.}
  \bibinfo{year}{2017}\natexlab{}.
\newblock \showarticletitle{DeepLab: Semantic Image Segmentation with Deep
  Convolutional Nets, Atrous Convolution, and Fully Connected CRFs}.
\newblock \bibinfo{journal}{{\em IEEE transactions on pattern analysis and
  machine intelligence\/}} (\bibinfo{year}{2017}).
\newblock


\bibitem[\protect\citeauthoryear{Chen, Li, and Tang}{Chen
  et~al\mbox{.}}{2013}]%
        {chen2013knn}
\bibfield{author}{\bibinfo{person}{Qifeng Chen}, \bibinfo{person}{Dingzeyu Li},
  {and} \bibinfo{person}{Chi-Keung Tang}.} \bibinfo{year}{2013}\natexlab{}.
\newblock \showarticletitle{KNN matting}.
\newblock \bibinfo{journal}{{\em IEEE transactions on pattern analysis and
  machine intelligence\/}} \bibinfo{volume}{35}, \bibinfo{number}{9}
  (\bibinfo{year}{2013}), \bibinfo{pages}{2175--2188}.
\newblock


\bibitem[\protect\citeauthoryear{Chen, Wang, Xu, He, and Lu}{Chen
  et~al\mbox{.}}{2016}]%
        {Chen2016AUM}
\bibfield{author}{\bibinfo{person}{Yingying Chen}, \bibinfo{person}{Jinqiao
  Wang}, \bibinfo{person}{Min Xu}, \bibinfo{person}{Xiangjian He}, {and}
  \bibinfo{person}{Hanqing Lu}.} \bibinfo{year}{2016}\natexlab{}.
\newblock \showarticletitle{A unified model sharing framework for moving object
  detection}.
\newblock \bibinfo{journal}{{\em Signal Processing\/}}  \bibinfo{volume}{124}
  (\bibinfo{year}{2016}), \bibinfo{pages}{72--80}.
\newblock


\bibitem[\protect\citeauthoryear{Cho, Tai, and Kweon}{Cho
  et~al\mbox{.}}{2016}]%
        {Cho2016Natural}
\bibfield{author}{\bibinfo{person}{Donghyeon Cho}, \bibinfo{person}{Yu~Wing
  Tai}, {and} \bibinfo{person}{Inso Kweon}.} \bibinfo{year}{2016}\natexlab{}.
\newblock \showarticletitle{Natural Image Matting Using Deep Convolutional
  Neural Networks}. In \bibinfo{booktitle}{{\em European Conference on Computer
  Vision (ECCV), 2016}}. \bibinfo{pages}{626--643}.
\newblock


\bibitem[\protect\citeauthoryear{Chuang, Curless, Salesin, and Szeliski}{Chuang
  et~al\mbox{.}}{2001}]%
        {chuang2001bayesian}
\bibfield{author}{\bibinfo{person}{Yung-Yu Chuang}, \bibinfo{person}{Brian
  Curless}, \bibinfo{person}{David~H Salesin}, {and} \bibinfo{person}{Richard
  Szeliski}.} \bibinfo{year}{2001}\natexlab{}.
\newblock \showarticletitle{A bayesian approach to digital matting}. In
  \bibinfo{booktitle}{{\em Computer Vision and Pattern Recognition (CVPR),
  2001. Proceedings of the 2001 IEEE Computer Society Conference on}},
  Vol.~\bibinfo{volume}{2}. IEEE, \bibinfo{pages}{II--II}.
\newblock


\bibitem[\protect\citeauthoryear{Gastal and Oliveira}{Gastal and
  Oliveira}{2010}]%
        {gastal2010shared}
\bibfield{author}{\bibinfo{person}{Eduardo~SL Gastal} {and}
  \bibinfo{person}{Manuel~M Oliveira}.} \bibinfo{year}{2010}\natexlab{}.
\newblock \showarticletitle{Shared sampling for real-time alpha matting}. In
  \bibinfo{booktitle}{{\em Computer Graphics Forum}},
  Vol.~\bibinfo{volume}{29}. Wiley Online Library, \bibinfo{pages}{575--584}.
\newblock


\bibitem[\protect\citeauthoryear{Handa, Patraucean, Badrinarayanan, Stent, and
  Cipolla}{Handa et~al\mbox{.}}{2015}]%
        {Handa2015SceneNetUR}
\bibfield{author}{\bibinfo{person}{Ankur Handa}, \bibinfo{person}{Viorica
  Patraucean}, \bibinfo{person}{Vijay Badrinarayanan}, \bibinfo{person}{Simon
  Stent}, {and} \bibinfo{person}{Roberto Cipolla}.}
  \bibinfo{year}{2015}\natexlab{}.
\newblock \showarticletitle{SceneNet: Understanding Real World Indoor Scenes
  With Synthetic Data}.
\newblock \bibinfo{journal}{{\em CoRR\/}}  \bibinfo{volume}{abs/1511.07041}
  (\bibinfo{year}{2015}).
\newblock


\bibitem[\protect\citeauthoryear{He, Sun, and Tang}{He et~al\mbox{.}}{2010}]%
        {he2010fast}
\bibfield{author}{\bibinfo{person}{Kaiming He}, \bibinfo{person}{Jian Sun},
  {and} \bibinfo{person}{Xiaoou Tang}.} \bibinfo{year}{2010}\natexlab{}.
\newblock \showarticletitle{Fast matting using large kernel matting laplacian
  matrices}. In \bibinfo{booktitle}{{\em Computer Vision and Pattern
  Recognition (CVPR), 2010 IEEE Conference on}}. IEEE,
  \bibinfo{pages}{2165--2172}.
\newblock


\bibitem[\protect\citeauthoryear{He, Sun, and Tang}{He et~al\mbox{.}}{2013}]%
        {he2013guided}
\bibfield{author}{\bibinfo{person}{Kaiming He}, \bibinfo{person}{Jian Sun},
  {and} \bibinfo{person}{Xiaoou Tang}.} \bibinfo{year}{2013}\natexlab{}.
\newblock \showarticletitle{Guided image filtering}.
\newblock \bibinfo{journal}{{\em IEEE transactions on pattern analysis and
  machine intelligence\/}} \bibinfo{volume}{35}, \bibinfo{number}{6}
  (\bibinfo{year}{2013}), \bibinfo{pages}{1397--1409}.
\newblock


\bibitem[\protect\citeauthoryear{Hu and Ramanan}{Hu and Ramanan}{2017}]%
        {Hu2016FindingTF}
\bibfield{author}{\bibinfo{person}{Peiyun Hu} {and} \bibinfo{person}{Deva
  Ramanan}.} \bibinfo{year}{2017}\natexlab{}.
\newblock \showarticletitle{Finding Tiny Faces}. In \bibinfo{booktitle}{{\em
  Computer Vision and Pattern Recognition (CVPR), 2017}}.
\newblock


\bibitem[\protect\citeauthoryear{Huang, Liu, Weinberger, and van~der
  Maaten}{Huang et~al\mbox{.}}{2017}]%
        {huang2016densely}
\bibfield{author}{\bibinfo{person}{Gao Huang}, \bibinfo{person}{Zhuang Liu},
  \bibinfo{person}{Kilian~Q Weinberger}, {and} \bibinfo{person}{Laurens van~der
  Maaten}.} \bibinfo{year}{2017}\natexlab{}.
\newblock \showarticletitle{Densely connected convolutional networks}. In
  \bibinfo{booktitle}{{\em Computer Vision and Pattern Recognition (CVPR),
  2017}}.
\newblock


\bibitem[\protect\citeauthoryear{J{\'{e}}gou, Drozdzal, V{\'{a}}zquez, Romero,
  and Bengio}{J{\'{e}}gou et~al\mbox{.}}{2017}]%
        {Jgou2016TheOH}
\bibfield{author}{\bibinfo{person}{Simon J{\'{e}}gou}, \bibinfo{person}{Michal
  Drozdzal}, \bibinfo{person}{David V{\'{a}}zquez}, \bibinfo{person}{Adriana
  Romero}, {and} \bibinfo{person}{Yoshua Bengio}.}
  \bibinfo{year}{2017}\natexlab{}.
\newblock \showarticletitle{The One Hundred Layers Tiramisu: Fully
  Convolutional DenseNets for Semantic Segmentation}. In
  \bibinfo{booktitle}{{\em Workshop on Computer Vision in Vehicle Technology
  CVPR, 2017}}.
\newblock


\bibitem[\protect\citeauthoryear{Jia, Shelhamer, Donahue, Karayev, Long,
  Girshick, Guadarrama, and Darrell}{Jia et~al\mbox{.}}{2014}]%
        {jia2014caffe}
\bibfield{author}{\bibinfo{person}{Yangqing Jia}, \bibinfo{person}{Evan
  Shelhamer}, \bibinfo{person}{Jeff Donahue}, \bibinfo{person}{Sergey Karayev},
  \bibinfo{person}{Jonathan Long}, \bibinfo{person}{Ross Girshick},
  \bibinfo{person}{Sergio Guadarrama}, {and} \bibinfo{person}{Trevor Darrell}.}
  \bibinfo{year}{2014}\natexlab{}.
\newblock \showarticletitle{Caffe: Convolutional architecture for fast feature
  embedding}. In \bibinfo{booktitle}{{\em Proceedings of the 22nd ACM
  international conference on Multimedia}}. ACM, \bibinfo{pages}{675--678}.
\newblock


\bibitem[\protect\citeauthoryear{Latifi~Oskouei, Golestani, Hashemi, and
  Ghiasi}{Latifi~Oskouei et~al\mbox{.}}{2016}]%
        {cnndroid2016}
\bibfield{author}{\bibinfo{person}{Seyyed~Salar Latifi~Oskouei},
  \bibinfo{person}{Hossein Golestani}, \bibinfo{person}{Matin Hashemi}, {and}
  \bibinfo{person}{Soheil Ghiasi}.} \bibinfo{year}{2016}\natexlab{}.
\newblock \showarticletitle{CNNdroid: GPU-Accelerated Execution of Trained Deep
  Convolutional Neural Networks on Android}. In \bibinfo{booktitle}{{\em
  Proceedings of the 2016 ACM on Multimedia Conference}}.
  \bibinfo{pages}{1201--1205}.
\newblock


\bibitem[\protect\citeauthoryear{Levin, Lischinski, and Weiss}{Levin
  et~al\mbox{.}}{2008}]%
        {levin2008closed}
\bibfield{author}{\bibinfo{person}{Anat Levin}, \bibinfo{person}{Dani
  Lischinski}, {and} \bibinfo{person}{Yair Weiss}.}
  \bibinfo{year}{2008}\natexlab{}.
\newblock \showarticletitle{A closed-form solution to natural image matting}.
\newblock \bibinfo{journal}{{\em IEEE Transactions on Pattern Analysis and
  Machine Intelligence\/}} \bibinfo{volume}{30}, \bibinfo{number}{2}
  (\bibinfo{year}{2008}), \bibinfo{pages}{228--242}.
\newblock


\bibitem[\protect\citeauthoryear{Liu, Anguelov, Erhan, Szegedy, Reed, Fu, and
  Berg}{Liu et~al\mbox{.}}{2016}]%
        {Liu2016SSDSS}
\bibfield{author}{\bibinfo{person}{Wei Liu}, \bibinfo{person}{Dragomir
  Anguelov}, \bibinfo{person}{Dumitru Erhan}, \bibinfo{person}{Christian
  Szegedy}, \bibinfo{person}{Scott~E. Reed}, \bibinfo{person}{Cheng-Yang Fu},
  {and} \bibinfo{person}{Alexander~C. Berg}.} \bibinfo{year}{2016}\natexlab{}.
\newblock \showarticletitle{SSD: Single Shot MultiBox Detector}. In
  \bibinfo{booktitle}{{\em ECCV}}.
\newblock


\bibitem[\protect\citeauthoryear{Long, Shelhamer, and Darrell}{Long
  et~al\mbox{.}}{2015}]%
        {long2015fully}
\bibfield{author}{\bibinfo{person}{Jonathan Long}, \bibinfo{person}{Evan
  Shelhamer}, {and} \bibinfo{person}{Trevor Darrell}.}
  \bibinfo{year}{2015}\natexlab{}.
\newblock \showarticletitle{Fully convolutional networks for semantic
  segmentation}. In \bibinfo{booktitle}{{\em Proceedings of the IEEE Conference
  on Computer Vision and Pattern Recognition}}. \bibinfo{pages}{3431--3440}.
\newblock


\bibitem[\protect\citeauthoryear{Paszke, Chaurasia, Kim, and
  Culurciello}{Paszke et~al\mbox{.}}{2016}]%
        {paszke2016enet}
\bibfield{author}{\bibinfo{person}{Adam Paszke}, \bibinfo{person}{Abhishek
  Chaurasia}, \bibinfo{person}{Sangpil Kim}, {and} \bibinfo{person}{Eugenio
  Culurciello}.} \bibinfo{year}{2016}\natexlab{}.
\newblock \showarticletitle{Enet: A deep neural network architecture for
  real-time semantic segmentation}.
\newblock \bibinfo{journal}{{\em arXiv preprint arXiv:1606.02147\/}}
  (\bibinfo{year}{2016}).
\newblock


\bibitem[\protect\citeauthoryear{Peng, Zhang, Yu, Luo, and Sun}{Peng
  et~al\mbox{.}}{2017}]%
        {Peng2017LargeKM}
\bibfield{author}{\bibinfo{person}{Chao Peng}, \bibinfo{person}{Xiangyu Zhang},
  \bibinfo{person}{Gang Yu}, \bibinfo{person}{Guiming Luo}, {and}
  \bibinfo{person}{Jian Sun}.} \bibinfo{year}{2017}\natexlab{}.
\newblock \showarticletitle{Large Kernel Matters - Improve Semantic
  Segmentation by Global Convolutional Network}. In \bibinfo{booktitle}{{\em
  Computer Vision and Pattern Recognition (CVPR), 2017}}.
\newblock


\bibitem[\protect\citeauthoryear{Qi, Chi, Peter, and Ho}{Qi
  et~al\mbox{.}}{2016}]%
        {Qi2016CASAIRCA}
\bibfield{author}{\bibinfo{person}{Shaoyu Qi}, \bibinfo{person}{Yu-Tseh Chi},
  \bibinfo{person}{Adrian~M. Peter}, {and} \bibinfo{person}{Jeffrey Ho}.}
  \bibinfo{year}{2016}\natexlab{}.
\newblock \showarticletitle{CASAIR: Content and Shape-Aware Image Retargeting
  and Its Applications}.
\newblock \bibinfo{journal}{{\em IEEE Transactions on Image Processing\/}}
  \bibinfo{volume}{25} (\bibinfo{year}{2016}), \bibinfo{pages}{2222--2232}.
\newblock


\bibitem[\protect\citeauthoryear{Qin, Yan, Li, and Hu}{Qin
  et~al\mbox{.}}{2016}]%
        {Qin2016JointTO}
\bibfield{author}{\bibinfo{person}{Hongwei Qin}, \bibinfo{person}{Junjie Yan},
  \bibinfo{person}{Xiu Li}, {and} \bibinfo{person}{Xiaolin Hu}.}
  \bibinfo{year}{2016}\natexlab{}.
\newblock \showarticletitle{Joint Training of Cascaded CNN for Face Detection}.
\newblock \bibinfo{journal}{{\em 2016 IEEE Conference on Computer Vision and
  Pattern Recognition (CVPR)\/}} (\bibinfo{year}{2016}),
  \bibinfo{pages}{3456--3465}.
\newblock


\bibitem[\protect\citeauthoryear{Redmon and Farhadi}{Redmon and
  Farhadi}{2017}]%
        {Redmon2016YOLO9000BF}
\bibfield{author}{\bibinfo{person}{Joseph Redmon} {and} \bibinfo{person}{Ali
  Farhadi}.} \bibinfo{year}{2017}\natexlab{}.
\newblock \showarticletitle{YOLO9000: Better, Faster, Stronger}. In
  \bibinfo{booktitle}{{\em Computer Vision and Pattern Recognition (CVPR),
  2017}}.
\newblock


\bibitem[\protect\citeauthoryear{Ren, He, Girshick, and Sun}{Ren
  et~al\mbox{.}}{2015}]%
        {Ren2015FasterRT}
\bibfield{author}{\bibinfo{person}{Shaoqing Ren}, \bibinfo{person}{Kaiming He},
  \bibinfo{person}{Ross Girshick}, {and} \bibinfo{person}{Jian Sun}.}
  \bibinfo{year}{2015}\natexlab{}.
\newblock \showarticletitle{Faster R-CNN: Towards Real-Time Object Detection
  with Region Proposal Networks}.
\newblock \bibinfo{journal}{{\em IEEE Transactions on Pattern Analysis and
  Machine Intelligence\/}}  \bibinfo{volume}{39} (\bibinfo{year}{2015}),
  \bibinfo{pages}{1137--1149}.
\newblock


\bibitem[\protect\citeauthoryear{Schroff, Kalenichenko, and Philbin}{Schroff
  et~al\mbox{.}}{2015}]%
        {Schroff2015FaceNetAU}
\bibfield{author}{\bibinfo{person}{Florian Schroff}, \bibinfo{person}{Dmitry
  Kalenichenko}, {and} \bibinfo{person}{James Philbin}.}
  \bibinfo{year}{2015}\natexlab{}.
\newblock \showarticletitle{FaceNet: A unified embedding for face recognition
  and clustering}.
\newblock \bibinfo{journal}{{\em 2015 IEEE Conference on Computer Vision and
  Pattern Recognition (CVPR)\/}} (\bibinfo{year}{2015}),
  \bibinfo{pages}{815--823}.
\newblock


\bibitem[\protect\citeauthoryear{Shen, Hertzmann, Jia, Paris, Price, Shechtman,
  and Sachs}{Shen et~al\mbox{.}}{2016a}]%
        {shen2016automatic}
\bibfield{author}{\bibinfo{person}{Xiaoyong Shen}, \bibinfo{person}{Aaron
  Hertzmann}, \bibinfo{person}{Jiaya Jia}, \bibinfo{person}{Sylvain Paris},
  \bibinfo{person}{Brian Price}, \bibinfo{person}{Eli Shechtman}, {and}
  \bibinfo{person}{Ian Sachs}.} \bibinfo{year}{2016}\natexlab{a}.
\newblock \showarticletitle{Automatic portrait segmentation for image
  stylization}. In \bibinfo{booktitle}{{\em Computer Graphics Forum}},
  Vol.~\bibinfo{volume}{35}. Wiley Online Library, \bibinfo{pages}{93--102}.
\newblock


\bibitem[\protect\citeauthoryear{Shen, Tao, Gao, Zhou, and Jia}{Shen
  et~al\mbox{.}}{2016b}]%
        {shen2016deep}
\bibfield{author}{\bibinfo{person}{Xiaoyong Shen}, \bibinfo{person}{Xin Tao},
  \bibinfo{person}{Hongyun Gao}, \bibinfo{person}{Chao Zhou}, {and}
  \bibinfo{person}{Jiaya Jia}.} \bibinfo{year}{2016}\natexlab{b}.
\newblock \showarticletitle{Deep Automatic Portrait Matting}. In
  \bibinfo{booktitle}{{\em European Conference on Computer Vision}}. Springer,
  \bibinfo{pages}{92--107}.
\newblock


\bibitem[\protect\citeauthoryear{Sun, Jia, Tang, and Shum}{Sun
  et~al\mbox{.}}{2004}]%
        {sun2004poisson}
\bibfield{author}{\bibinfo{person}{Jian Sun}, \bibinfo{person}{Jiaya Jia},
  \bibinfo{person}{Chi-Keung Tang}, {and} \bibinfo{person}{Heung-Yeung Shum}.}
  \bibinfo{year}{2004}\natexlab{}.
\newblock \showarticletitle{Poisson matting}. In \bibinfo{booktitle}{{\em ACM
  Transactions on Graphics (ToG)}}, Vol.~\bibinfo{volume}{23}. ACM,
  \bibinfo{pages}{315--321}.
\newblock


\bibitem[\protect\citeauthoryear{Szegedy, Ioffe, Vanhoucke, and Alemi}{Szegedy
  et~al\mbox{.}}{2016}]%
        {szegedy2016inception}
\bibfield{author}{\bibinfo{person}{Christian Szegedy}, \bibinfo{person}{Sergey
  Ioffe}, \bibinfo{person}{Vincent Vanhoucke}, {and} \bibinfo{person}{Alex~A.
  Alemi}.} \bibinfo{year}{2016}\natexlab{}.
\newblock \showarticletitle{Inception-v4, Inception-ResNet and the Impact of
  Residual Connections on Learning}. In \bibinfo{booktitle}{{\em ICLR 2016
  Workshop}}.
\newblock


\bibitem[\protect\citeauthoryear{Wang, Duan, Liu, Lu, and Jin}{Wang
  et~al\mbox{.}}{2008}]%
        {Wang2008AMS}
\bibfield{author}{\bibinfo{person}{Jinqiao Wang}, \bibinfo{person}{Ling-Yu
  Duan}, \bibinfo{person}{Qingshan Liu}, \bibinfo{person}{Hanqing Lu}, {and}
  \bibinfo{person}{Jesse~S. Jin}.} \bibinfo{year}{2008}\natexlab{}.
\newblock \showarticletitle{A Multimodal Scheme for Program Segmentation and
  Representation in Broadcast Video Streams}.
\newblock \bibinfo{journal}{{\em IEEE Trans. Multimedia\/}}
  \bibinfo{volume}{10} (\bibinfo{year}{2008}), \bibinfo{pages}{393--408}.
\newblock


\bibitem[\protect\citeauthoryear{Wang, Fu, Lu, and Ma}{Wang
  et~al\mbox{.}}{2014}]%
        {Wang2014BilayerST}
\bibfield{author}{\bibinfo{person}{Jinqiao Wang}, \bibinfo{person}{Wei Fu},
  \bibinfo{person}{Hanqing Lu}, {and} \bibinfo{person}{Songde Ma}.}
  \bibinfo{year}{2014}\natexlab{}.
\newblock \showarticletitle{Bilayer Sparse Topic Model for Scene Analysis in
  Imbalanced Surveillance Videos}.
\newblock \bibinfo{journal}{{\em IEEE Transactions on Image Processing\/}}
  \bibinfo{volume}{23} (\bibinfo{year}{2014}), \bibinfo{pages}{5198--5208}.
\newblock


\bibitem[\protect\citeauthoryear{Wang, Qu, Chen, Mei, Xu, Zhang, and Lu}{Wang
  et~al\mbox{.}}{2016}]%
        {Wang2016AdaptiveCC}
\bibfield{author}{\bibinfo{person}{Jinqiao Wang}, \bibinfo{person}{Zhan Qu},
  \bibinfo{person}{Yingying Chen}, \bibinfo{person}{Tao Mei},
  \bibinfo{person}{Min Xu}, \bibinfo{person}{La Zhang}, {and}
  \bibinfo{person}{Hanqing Lu}.} \bibinfo{year}{2016}\natexlab{}.
\newblock \showarticletitle{Adaptive Content Condensation Based on Grid
  Optimization for Thumbnail Image Generation}.
\newblock \bibinfo{journal}{{\em IEEE Trans. Circuits Syst. Video Techn.\/}}
  \bibinfo{volume}{26} (\bibinfo{year}{2016}), \bibinfo{pages}{2079--2092}.
\newblock


\bibitem[\protect\citeauthoryear{Xu, Price, Cohen, and Huang}{Xu
  et~al\mbox{.}}{2017}]%
        {xu2017deep}
\bibfield{author}{\bibinfo{person}{Ning Xu}, \bibinfo{person}{Brian Price},
  \bibinfo{person}{Scott Cohen}, {and} \bibinfo{person}{Thomas Huang}.}
  \bibinfo{year}{2017}\natexlab{}.
\newblock \showarticletitle{Deep Image Matting}. In \bibinfo{booktitle}{{\em
  Computer Vision and Pattern Recognition (CVPR), 2017}}.
\newblock


\bibitem[\protect\citeauthoryear{Yu and Koltun}{Yu and Koltun}{2016}]%
        {yu2015multi}
\bibfield{author}{\bibinfo{person}{Fisher Yu} {and} \bibinfo{person}{Vladlen
  Koltun}.} \bibinfo{year}{2016}\natexlab{}.
\newblock \showarticletitle{Multi-scale context aggregation by dilated
  convolutions}. In \bibinfo{booktitle}{{\em International Conference on
  Learning Representations (ICLR), 2016}}.
\newblock


\bibitem[\protect\citeauthoryear{Zhao, Shi, Qi, Wang, and Jia}{Zhao
  et~al\mbox{.}}{2017}]%
        {zhao2016pyramid}
\bibfield{author}{\bibinfo{person}{Hengshuang Zhao}, \bibinfo{person}{Jianping
  Shi}, \bibinfo{person}{Xiaojuan Qi}, \bibinfo{person}{Xiaogang Wang}, {and}
  \bibinfo{person}{Jiaya Jia}.} \bibinfo{year}{2017}\natexlab{}.
\newblock \showarticletitle{Pyramid Scene Parsing Network}. In
  \bibinfo{booktitle}{{\em Computer Vision and Pattern Recognition (CVPR),
  2017}}.
\newblock


\bibitem[\protect\citeauthoryear{Zhou, Zhao, Puig, Fidler, Barriuso, and
  Torralba}{Zhou et~al\mbox{.}}{2016}]%
        {Zhou2016SemanticUO}
\bibfield{author}{\bibinfo{person}{Bolei Zhou}, \bibinfo{person}{Hang Zhao},
  \bibinfo{person}{Xavier Puig}, \bibinfo{person}{Sanja Fidler},
  \bibinfo{person}{Adela Barriuso}, {and} \bibinfo{person}{Antonio Torralba}.}
  \bibinfo{year}{2016}\natexlab{}.
\newblock \showarticletitle{Semantic Understanding of Scenes through the ADE20K
  Dataset}.
\newblock \bibinfo{journal}{{\em CoRR\/}}  \bibinfo{volume}{abs/1608.05442}
  (\bibinfo{year}{2016}).
\newblock


\end{thebibliography}

\end{document}